\documentclass[accepted]{uai2023} 

\usepackage[american]{babel}
\usepackage{multirow}

\usepackage{natbib} 
\usepackage{mathtools} 
\usepackage{booktabs} 
\usepackage{tikz} 

\newtheorem{theorem}{Theorem}
\newtheorem{definition}{Definition}

\newtheorem{lemma}{Lemma}
\newtheorem{proof}{Proof}

\usepackage{cancel}

\usepackage{amsmath,amssymb}
\usepackage{mathrsfs}  
\usepackage{dsfont}

\newtheorem{remark}{Remark}

\usepackage{csquotes}
\usepackage{xr} 
\DeclareMathOperator{\und}{\;\&\;}

\DeclareSymbolFont{rsfs}{U}{rsfs}{m}{n}
\DeclareSymbolFontAlphabet{\mathrsfs}{rsfs}

\newcommand{\Pcal}{\mathcal{P}}
\newcommand{\Sscr}{\mathscr{S}}

\title{A note on the connectedness property of union-free generic sets of partial orders}

\author[1]{\href{mailto:georg.schollmeyer@stat.uni-muenchen.de}{Georg~Schollmeyer}{}}
\author[1]{Hannah~Blocher}

\affil[1]{%
    Department of Statistics\\
    Ludwig-Maximilians-Universität\\
    Munich, Bavaria, Germany
}

\begin{document} 

\onecolumn 
\maketitle

{\bfseries{Abstract}}
  This short note describes and proves a connectedness property which was introduced in \cite{Blocher_isipta} in the context of data depth functions for partial orders. The connectedness property gives a structural insight into union-free generic sets. These sets, presented in \cite{Blocher_isipta}, are defined by using a closure operator on the set of all partial orders which naturally appears within the theory of formal concept analysis. In the language of formal concept analysis, the property of connectedness can be vividly proven. However, since within \cite{Blocher_isipta} we did not discuss formal concept analysis, we outsourced the proof to this note.
  
  


  \section{Introduction}

    \textcolor{red}{\textbf{Corrigendum}}
    
  \noindent\fbox{%
    \parbox{\textwidth}{%
        The connectedness proof given here is not valid. A corrected proof can be found in Hannah Blocher, Georg Schollmeyer, Malte Nalenz, and Christoph Jansen (2023): Comparing Machine Learning Algorithms by Union-Free Generic Depth, ArXiv report \url{https://arxiv.org/abs/2312.12839} (viewed December 21, 2023). \vspace{1em}
        
        The error occurs at the yellow box below. Consider the following counterexample: Let $M = \{a,b,a_1, b_1, c_1\}$. Consider the posets $p_1 = \{(a,b), (a_1, c_1)\}, p_2 = \{(a_1, b_1)\}$, and $p_3 = \{(b_1, c_1)\}$. Then $\{p_1, p_2, p_3\}$ is union-free and generic, since $q = \{(a,b), (a_1, b_1), (b_1, c_1), (a_1, c_1)\}$ lies only in the closure of $p_1, p_2$ and $p_3$, but not in a proper subset, see Lemma 1. Let this $q$ be the $q$ of case i) on page 6.
        
        The problem we did not consider is that not every poset that has an edge as described there can be taken out. For example, in this case the edge $(a,b)$ given by $p_1$ satisfies all the properties in (*), but we cannot remove the partial order $p_1$ because we need the edge $(a_1, c_1)$. Here we also have to consider that we can only remove a poset $p$ if and only if it is not needed for any transitive structure not stemming from $p$. Of course, one can construct a poset $q$ such that every poset in the union-free and generic set is represented by transitivity in $q$, but then there is another $\tilde{q} \subseteq q$ where at least one poset in the union-free and generic set satisfies the necessary property to be removed. For more details, see Hannah Blocher, Georg Schollmeyer, Malte Nalenz, and Christoph Jansen (2023): Comparing Machine Learning Algorithms by Union-Free Generic Depth, ArXiv report \url{https://arxiv.org/abs/2312.12839} (viewed December 21, 2023).
    }%
}

\vspace{1em}

In \cite{Blocher_isipta} we introduced a depth function, the ufg-depth, for the set of partial orders and with this we gave a notion of centrality for partial orders. The presented ufg-depth is based on sets which are \textit{union-free} and \textit{generic} elemnts of a closure system. To obtain the closure operator on the set of partial orders, we used the theory of formal concept analysis (FCA, see below). Formal concept analysis can be used very broadly to represent many kinds of non-standard data in what is called a formal context. From this formal context, one always obtains a closure operator that characterizes the underlying data set, in our case the observed partial orders.%



Our aim in this note is now to prove the \textit{connectedness} property of these union-free generic sets. This property was given in \cite{Blocher_isipta} and used to improve the implementation of ufg-depth. The proof uses that the corresponding closure operator is given by the formal context introduced in \cite{bsj2022}. Since in \cite{Blocher_isipta} all constructions and considerations were made without reference to formal concept analysis, and the proof of the connectedness property is most easily presented in the language of formal concept analysis, we decided to outsource the proof to this short note. 

Therefore, the structure of this note is as follows: Section~\ref{Intro_FCA} gives an introduction to partial orders and the main ideas of formal concept analysis. Afterwards, we present the formal context given by \cite{bsj2022} and the resulting closure system. Furthermore, we define the union-free generic sets as in \cite{Blocher_isipta}. Section~\ref{sec: proof} states and proves the connectedness property.

\section{Preliminaries: Partially ordered sets and Formal concept analysis}\label{Intro_FCA}
We start by giving some basic definitions of order theory and formal concept analysis. We refer to \cite[chapters 0-2]{ganter2012formal} or \cite{priss2006formal} for a far more extensive and detailed introduction.
\begin{definition}[Partially ordered set]

Let $M$ be a fixed set. Then a \textbf{partially ordered set (poset)} on $M$ is a binary relation $p\subseteq M\times M$ that is 
\begin{enumerate}
    \item reflexive: $\quad \forall y \in M: (y,y) \in p$,
    \item transitive: $\quad \forall y_1,y_2,y_3 \in M: (y_1,y_2) \in p , (y_2,y_3) \in p \Longrightarrow (y_1,y_3) \in p$ and
    \item antisymmetric: $\quad \forall y_1,y_2 \in M: (y_1,y_2)\in p, y_1 \neq y_2 \Longrightarrow (y_2,y_1) \notin p$.
\end{enumerate}

For a fixed set $M$ we denote with $\mathcal{P}$ the set of all partial orders on $M$.

\end{definition}

Later, see Section~\ref{sec: ufg sets def} we represent the set of partial orders within the theory of formal concept analysis. This is done via a formal context:
\begin{definition}[Formal context, formal concept and associated closure operator]
A \textbf{formal context} is a triple  $\mathbb{K}=(G,M,I)$ where $G$ is a set of \textbf{objects}, $M$ is a set of \textbf{attributes} and $I\subseteq G \times M$ is a binary relation between the objects and the attributes with the interpretation $(g,m) \in I$ iff object $g$ has attribute $m$. If $(g,m)\in I$ we also use infix notation and write $gIm$. 
Given a formal context $\mathbb{K}=(G,MI,)$ we define the following operators:
\begin{align*}
\Phi: 2^M \longrightarrow 2^G : &\quad B\mapsto \{g \in G \mid \forall m \in B: gIm\}\\
\Psi: 2^G \longrightarrow 2^M: &\quad  A \mapsto \{m \in M \mid \forall g \in A: gIm\}.
\end{align*}
Then we say that a pair $(A,B)$ with $A\subseteq G$ and $B\subseteq M$ is a formal concept if

$$\Psi(A)=B \und \Phi(B)=A .$$
Additionally, we define the operator
$$\gamma := \Phi \circ \Psi.$$
This operator is a \textbf{closure operator} on $G$. This means that $\gamma$ is:

\begin{enumerate}
    \item \textbf{extensive}: $\forall A\subseteq G: A\subseteq \gamma(A)$,
    \item \textbf{isotone}: $\forall A,B \subseteq G: A\subseteq B \Longrightarrow \gamma(A) \subseteq \gamma(B)$ and
    \item \textbf{idempotent}: $\forall A\subseteq G: \gamma(\gamma(A))=\gamma(A).$
\end{enumerate}
The sets $\gamma(A)$ with $A\subseteq G$ are called the \textbf{hulls} of $\gamma$.
\end{definition}

\begin{remark}
The family $\{\gamma(A) \mid A\subseteq G\}$ of all hulls of $\gamma$ is a closure-system. This means that it contains the set $G$ and it is closed under arbitrary intersections.
Note that there is a one-to-one relation between a closure operator and a closure system and that both from the closure operator, as well as from the corresponding closure system associated to a formal context, one can get back the whole context. In this sense, $\gamma$ (or the hulls $\{\gamma(A)\mid A\subseteq G\}$) represents the formal context.
\end{remark}

After defining closure operators/systems and formal contexts and observing the one-to-one correspondence, we can go a step further. These two mathematical concepts have also a one-to-one correspondence to the family of implications. Generally speaking, implications describe the dependencies between the objects $G$.

\begin{definition}[Formal implication]
Every closure operator $\gamma$  on $G$ can be described by all valid formal implications: A formal implication is a pair $(Y,Z)$ of subsets of $G$. 
We say that an implication $(Y,Z)$ is valid w.r.t. a closure operator $\gamma$  if $\gamma(Y) \supseteq \gamma(Z)$ and denote this by $Y\longrightarrow Z$.
The first component of a formal implication is also called the {\bfseries{premise}} or the {\bfseries{antecedent}} and the second component is also called {\bfseries{conclusion }} or the {\bfseries{consequent}} of the formal implication.
\end{definition}

In the context of statistical data analysis one often has data that are not binary but for example categorical or, as in our case, every data point is one partial order (on an underlying set). To analyze such data with methods of formal concept analysis one can use the technique of {\bfseries{conceptual scaling}} (cf. \citep[p.36-45]{ganter2012formal}) to fit such data into a binary setting: For example a categorical variable with the possible values in $\{1,\ldots ,K\}$ one can introduce the $K$ attributes ``$=1$''$, \ldots ,$``$=K$'' and say that an object $g$ has attribute ``$=i$'' if the value of $K$ equals $i$. In a similar way, for an ordinal variable with possible values $\{1<2<\ldots < K\}$ we can introduce the attributes ``$\leq 1$'',``$\leq 2$'',$\ldots$, ``$\leq K$'' and say that object $g$ has attribute ``$\leq i$'' if the value of object $g$ is lower than or equal to $i$. One can also additionally introduce the attributes ``$\geq 1$'', $\ldots$, ``$\geq K$''. This concrete way of conceptually scaling an ordinal variable is called {\bfseries{interordinal scaling}}. For the case of data that are partial orders, the authors are not aware of any literature that uses formal concept analysis together with a conceptual scaling. Therefore, in \cite{bsj2022} we introduced a conceptual scaling that we think is adequate for partially ordered data (see the discussion therein).


\section{Generic and union-free sets}\label{sec: ufg sets def}


In \cite{bsj2022} we defined the formal context $\mathbb{K}=(G,M,I)$ with $ G =\Pcal$ the set of all partial orders on a fixed finite set $\mathcal{X}=\{x_1,\ldots,x_N\}$. As an appropriate conceptual scaling we use as attributes the set $M$ defined by 
\begin{align*}
	\hspace{-0.15cm}M &:= \underbrace{\{\mbox{\enquote{$x_i \leq x_j$}}\mid i,j = 1, \ldots, N \text{, } i \neq j \}}_{=: M_{\leq}} \cup \underbrace{\{\mbox{\enquote{$x_i \not \leq x_j$}}\mid i,j = 1, \ldots, N \text{, } i \ne j \}}_{:= M_{\not\leq}}.
	\end{align*}

The binary relation $I$ is given by 
\begin{align}    
 (p\; , \;  \mbox{\enquote{$x_i \leq x_j$}} ) \in I:  & \iff (x_i,x_j) \in p & \mbox{ for }p \in \Pcal \mbox{ and } \mbox{\enquote{$x_i \leq x_j$}}  \in M_\leq \label{mleq}\\
 (p\;, \;  \mbox{\enquote{$x_i \not \leq x_j$}} ) \in I: &  \iff (x_i,x_j) \notin p & \mbox{ for } p \in \Pcal \mbox{ and } \mbox{\enquote{$x_i \not \leq x_j$}}  \in M_\nleq \label{mnleq}
 \end{align}



Now, according to Section~\ref{Intro_FCA}, the composition $\gamma \colon = \Phi \circ \Psi$ defines a closure operator. In the case of the above formal context, we get the following closure operator:
\begin{align*}
    \gamma\colon \begin{array}{l}
		2^{\Pcal} \to 2^{\Pcal}\\
		\{p_1, \ldots, p_k\} \mapsto \ \left\{p \in \Pcal \mid \bigcap\limits_{i = 1}^{k} p_i \subseteq p \subseteq \bigcup\limits_{i = 1}^{k} p_i \right\}.
	\end{array}	
\end{align*}
The part that each element of the hull must be a superset of the intersection follows from the $M_{\le}$ of the formal context. The second part, that each element must be a subset of the union, follows from the $M_{\not\le}$ attributes. Note that the intersection again defines a poset, while this does not hold for the union.

Now, we can define $\Sscr$ which is used to define the ufg depth. This set consists of elements of the closure system $\gamma(2^{\Pcal})$ which are generic and union-free. More precisely, we have $$\Sscr = \left\{P \subseteq\Pcal \mid \text{Condition } (C1) \text{ and } (C2) \text{ hold } \right\}$$ 
with Conditions (C1) and (C2) given by:
\begin{enumerate}
    \item[(C1)] $P \subsetneq \gamma(P),$
    \item[(C2)] There does not exist a family $(A_i)_{i \in \{1, \ldots, \ell\}}$ such that for all $i \in \{1, \ldots, \ell\}, \: A_i\subsetneq P$ and $\bigcup_{i \in \{1, \ldots, \ell\}} \gamma(A_i) = \gamma(P).$
\end{enumerate}

Condition (C1) states that $P$ is not a trivial set in the sense that it implies a strictly larger set than itself. Thus (C1) ensures that there exists a $\tilde{p} \in \Pcal\setminus P$ such that $P \to \{\tilde{p}\}$. Due to condition (C2), we only consider sets in  $\Sscr$ which are minimal. More precisely, Condition (C2) ensures that there is no suitable subset $\tilde{P} \subsetneq P$ such that $\gamma(P) = \gamma(\tilde{P})$ (just set $\ell = 1$ and $A_i = \tilde{P}$). Following \cite{bastide00}, we say that every element which has these two properties is called \textbf{generic}. Condition (C2) does more than guarantee that $P$ is minimal. It further ensures that each $P$ cannot be decomposed into proper subsets such that these subsets are sufficient to describe the hull of $P$. We call this property of having no such family of sets \textbf{union-free}. Thus $\Sscr$ consists of \textbf{union-free generic sets} or \textbf{ufg sets} for short.

\section{The connectedness property}\label{sec: proof}
In this section, we first state the claim of the main theorem. Then we give some preparatory definitions and lemmas. We end with the proof of the claim of the main theorem.

\subsection{Statement of the main theorem}

Now we state the main theorem. This theorem was used in \cite{Blocher_isipta} to make the enumeration of all ufg sets within the computation of the ufg depth compzutationally more efficient.

\begin{theorem}[Connectedness of ufg sets]\label{Connectedness of ufg sets}
The family of all ufg sets is connected in the following sense: 
Let $Q=\{p_1,p_2,p_3,\ldots,p_m\}$ be an ufg set of size $m\geq 3$. Then there exists a subset
$R\subsetneq Q$ of size $m-1$ that is an ufg set, too.
\end{theorem}

Theorem \ref{Connectedness of ufg sets} particularly implies that it is possible to enumerate all ufg sets by first enumerating all two-element ufg sets. Then one can recursively start with all currently given ufg sets and try to enlarge every ufg set by adding one partial order to get a new ufg set. If no partial order can be added to get a new ufg set, then one can stop with this set. With this, because of the connectedness property, one finds all ufg sets. Note that unfortunately one will possibly enumerate different ufg sets more than once. In the concrete implementation of the enumeration in \cite{Blocher_isipta} we avoid this by simply storing all already enumerated ufg sets in a list. Then, in every step of the recursive procedure, we look up if an ufg set candidate was already visited. In this case we stop the corresponding path within the enumearation procedure.

Note further that in the concrete implementation also other properties of the ufg sets are used:
\begin{enumerate}
    \item For an ufg set $Q= \{p_1,\ldots p_k\}$, every $q \in \gamma (\{p_1,\ldots, p_k\})$ cannot be a candidate for enlarging $Q$. 
    \item Also all partial orders that make an order in $p_i \in Q$ redundant (by removing all attributes (or more) that would be removed by $p_i$) are no candidate for enlarging $Q$, too.
\end{enumerate}

\subsection{Preparatory definitions and lemmas}

The next lemma states an equivalence definition of Condition (C1) and Condition (C2):
\begin{lemma}\label{lem:S_andere_def}
    Let $S \subseteq \mathcal{P}$. Then $S \in \Sscr$ if and only if there exists $q \in \gamma(S) \setminus S$ such that for all $x \in S, \: q \not\in \gamma(S\setminus \{x\})$.
\end{lemma}
\begin{proof}
    Let us first assume that $S \in \Sscr$. Then, due to Condition (C1), we know that $\gamma(S) \setminus S$ is nonempty. Since $\left( (S\setminus \{x\})_{x \in S} \right)$ is a family of proper subsets of $S$, we obtain by Condition (C2) that there must exist an element $q \in \gamma(S) \setminus S$ such that for all $x \in S, \: q \not\in \gamma(S\setminus \{x\})$.

    Conversely, suppose that there exists $q \in \gamma(S) \setminus S$ such that for all $x \in S, \: q \not \in \gamma(S\setminus \{x\})$. Condition (C1) follows immediately. To prove Condition (C2), let $(A_i)_{i \in \{1, \ldots, \ell\}}$ be such that for all $i \in \{1, \ldots, \ell\}, \: A_i\subsetneq S$ is arbitrary. Then for every $i \in \{1, \ldots, \ell\}$ there is an $x_i \in S$ with $x_i \not\in A_i$ (follows from $A_i \subsetneq S$). Since $\gamma$ is isotone, we know that $\gamma(A_i) \subseteq \gamma(S \setminus\{x_i\})$ and since $q \not\in \gamma(S \setminus\{x_i\})$, $q \not\in \gamma(A_i)$. Since this argument holds for every $i$, $q \not\in \bigcup_{i \in \{1, \ldots, \ell\}} \gamma(A_i)$. With $(A_i)_{i \in \{1, \ldots, \ell\}}$ arbitrarily chosen, the claim follows.
\end{proof}

Lemma~\ref{lem:S_andere_def} implies that a subset $S \subseteq \mathcal{P}$ is an element of $\Sscr$ if there is a poset $q \in \gamma(S)$ which can be represented only if every single $p_i \in S$ is contained. In the formal context of Section~\ref{sec: ufg sets def}, this means that for every $p_i \in S$ there must be an attribute that holds for every other $p_j \in S, \: p_i \neq p_j$, but not for $p_i$. This claim will be proved in Lemma~\ref{lem: dist_connect_ufg}, but first we need to define this property.

\begin{definition}
Let $S \subseteq G = (\mathcal{P})$ and $x \in S$. We define the \textit{distinguishing attributes} of $x$ w.r.t $S$ by
\begin{align*}
    \mathcal{D}_{x \in S} \colon = \{m \in M \mid x\not Im \text{ and for all } \hat{x} \in A \setminus \{x\}\colon  \hat{x}Im \}.
\end{align*}
Assume that $S\in \Sscr$. For $q \in \mathcal{P}$ we define the \textit{distinguishing attributes} of $x$ w.r.t $S$ and restricted to $q$ by
\begin{align*}
    \mathcal{D}^q_{x \in S} \colon = \{m \in M\setminus\Psi(q) \mid x\not Im \text{ and for all } \hat{x} \in A \setminus \{x\}\colon  xIm \}.
\end{align*}
Further, we will use the following notation 
\begin{align*}
    &\mathcal{D}^q_{\le} = \cup_{x \in S} \mathcal{D}_{x \in S} \cap M_{\le}\\
    &\mathcal{D}^q_{\not\le} = \cup_{x \in S} \mathcal{D}_{x \in S} \cap M_{\not\le}\\
\end{align*}
If it is clear which $q$ we are referring to, then we also write $\mathcal{D}_{\le}$ and $\mathcal{D}_{\not\le}$ instead of $\mathcal{D}^q_{\le}$ and $\mathcal{D}^q_{\not\le}$.
\end{definition}

The next lemma states how the distinguishing attributes can be used to verify if a set is union-free and generic.
    \begin{lemma}\label{lem: dist_connect_ufg}
        Let $S \subseteq G = \mathcal{P}$. Then $S \in \Sscr$ if and only if there exists a $q \in \gamma(S)$ such that for all $x \in S$, $\mathcal{D}^q_{x \in S} \neq \emptyset$.
    \end{lemma}
    \begin{proof}
        We prove that the right-hand side of Lemma~\ref{lem:S_andere_def} and the right-hand side of Lemma~\ref{lem: dist_connect_ufg} is equivalent. 
        
        First, we assume that the right-hand side of Lemma~\ref{lem:S_andere_def} is true. Thus, there exists a $q \in \Pcal$ such that $q \in \gamma(S)$ and for every $x \in S, \: q \not\in \gamma(S\setminus \{x\})$. Let $x \in S$ be arbitrary. Recall that $\gamma = \Phi \circ \Psi$ and therefore, there must exist an $m \in \Psi(S\setminus \{x\})$ such that $(q,m)\not\in I$, but for every $\tilde{x} \in S \setminus \{x\}, (\tilde{x}, m) \in I$. Since $q \in \gamma(S) \: (x,m) \not\in I$ must be true. Thus, $\mathcal{D}^q_{x \in S} \neq \emptyset$.
        
        Second, assume that the right-hand side of Lemma~\ref{lem: dist_connect_ufg} is true. Hence, there exists $q \in \Pcal$ such that for every $x \in S :\; D^q_{x \in S} \neq \emptyset$. This means that for every $x \in S$ there exists an $m \not\in \Psi(q)$ such that $(x,m) \not\in I$ and for all $\tilde{x} \in S \setminus \{x\}:\; (\tilde{x},m) \in I$. Thus, $q \not\in \gamma(S\setminus \{x\})$. This gives the right-hand side of Lemma~\ref{lem:S_andere_def}
    \end{proof}

Before we proof the main theorem in Section \ref{sec: proof} let us state some observations about the distinguishing elements.

{\bfseries{Remark 1:}} \label{lemma_characterizing_ufg}
Lemma~\ref{lem: dist_connect_ufg} can be reformulated by using implications as follows:
$Q=\{p_1,\ldots, p_k\}$ with $k\geq2$ is an ufg sets if and only if there exist attributes $m_1,\ldots, m_k \in M_{\leq} \cup M_{\nleq}$ 
and a partial order $q$ such that 
\begin{align}    
\{p_1,\ldots,p_k\}& \rightarrow \{q\} &\mbox{ and } \label{one}\\
\forall i,j \in \{1,\ldots,k\}:\;& p_i I m_j \iff i\neq j & \mbox{ and }\label{two}\\
\forall i \in\{1,\ldots,k\}:\;& q \cancel{I} m_i. \label{three}
\end{align}


Furthermore, note that there could be more than one $q \in \Pcal$ fulfilling the property on the right-hand side of Lemma~\ref{lem: dist_connect_ufg}. Additionally, for a poset $p_i$ there can be more than one distinguishing attribute (w.r.t. fixed $q$). In the sequel, we will collect in $D$ all distinguishing attributes w.r.t. some $p_i$ (and w.r.t. fixed $q$). Additionally, we will say that two distinguishing attributes $m$ and $\tilde{m}$ that distinguish the same order $p \in Q$ (i.e., $\forall i \in \{1,\ldots ,k\}: p_i I m \iff p_i I \tilde{m}$) are equivalent.  
\vspace{1em}

The next table illustrates the formal context for the set of partial orders and the distinguishing attributes of the ufg set $\{p_1, \ldots, p_k\}$. The object $q$ corresponds to the $q$ of the right-hand side of Lemma~\ref{lem: dist_connect_ufg}.
\begin{table}[ht]
        \centering
        \begin{tabular}{||l|cccc|ccc|c|}
            \hline
            & \multicolumn{4}{c|}{$D_\leq$}&\multicolumn{3}{c|}{$D_\nleq$}& rest\\
            
            & $m_1$ & $m_2$ & $m_3$& $\ldots$ & &$\ldots$ $m_{n-1}$ & $m_{k}$ & \\
            \hline
            $p_1$ &{\textcolor{red}{$\bigcirc$}}  & x & x&$\ldots$&$\ldots$ & x & x &\\
            $p_2$ & x &{\textcolor{red}{$\bigcirc$}}   & x &$\ldots$ &$\ldots$ &x & x &\\
            $p_3$ &x & x &{\textcolor{red}{$\bigcirc$}}  &$\ldots$ &$\ldots$&x&x&\\
            $\vdots$ &$\vdots$  & $\vdots$ & $\vdots$ &$\ddots$ &$\vdots$&$\vdots$  & $\vdots$&\\
            $\vdots$ &$\vdots$  & $\vdots$ & $\vdots$ &$\vdots$ &$\ddots$&$\vdots$  & $\vdots$&\\
            $p_{k-1}$ &x  &x & x& $\ldots$&$\ldots$ & {\textcolor{red}{$\bigcirc$}}  & x &\\
            $p_{k}$ & x & x &x  &$\ldots$&$\ldots$&x &{\textcolor{red}{$\bigcirc$}}   &\\
            \hline
            $q$ & {\textcolor{red}{$\bigcirc$}} & {\textcolor{red}{$\bigcirc$}} & {\textcolor{red}{$\bigcirc$}} &$\ldots$&$\ldots$& {\textcolor{red}{$\bigcirc$}} &{\textcolor{red}{$\bigcirc$}}&\\
            \hline
        \end{tabular}
    \end{table}

\subsection{Proof of the main theorem}
\begin{proof}[of Theorem 1]

Let $Q=\{p_1, \ldots, p_m\}$ be an ufg set of size $m\geq 3$ and let $D=D_{\leq} \cup D_{\nleq}$ be the set of distinguishing attributes w.r.t. some partial order $q$. In the following, we will show the claim by showing that we can always modify (or replace) the order $q$ to (by) another order $\tilde{q}$ that shows via the characterizing properties (\ref{one})-\ref{three}) that for an appropriately chosen order $p_i$ the corresponding proper subset $Q \backslash\{p_i\}$ of size $m-1$ is again an ufg set. Concretely, we distinguish two cases (where case i) is the more difficult case):\\

\textbf{Case} $\boldsymbol{i):}$ As long as $D_{\nleq}$ is not empty it is possible to select some distinguishing attribute $m_i \in D_{\nleq}$ and 'remove' the corresponding $p_i$ (such that $m_i$ and all possible other distinguishing attributes w.r.t. $p_i$ become irrelevant for the reduced set $Q\backslash \{p_i\}$) and modify $q$ to $\tilde{q}$ to get the subset $Q\backslash\{p_i\}$ from which we can show with the help of $\tilde{q}$ that this subset of size $m-1$ is also an ufg set. 

{\bfseries{Remark:}}  In this case we have to deal with the possible difficulty that for a partial order $p_i$ there exists more than one distinguishing attributes in $D_{\nleq}$. In this case, one has to repeat the recursive procedure described below until one did firstly remove all distinguishing attributes of only one order $p_i$ (such that all other orders $p_j$ will not become redundant). In case $i)$ it will be important to remove distinguishing attributes by modifying $q$ recursively to other orders $q_1\ldots, q_k=:\tilde{q}$ in such a way that the final order $q_k$ (and all other orders $q_l$) are in fact partial orders and in such a way that one removes not all distinguishing attributes of more than one order. (This is needed to make sure that the resulting subset of partial orders is not redundant in implying the resulting $\tilde{q}$.)\\

If $D_{\nleq}$ is empty, we are in \\

\textbf{Case} $\boldsymbol{ii):}$ Let $D_{\nleq} = \emptyset$. In this case we can remove an arbitrary $m_i \in D_{\leq}$ along with the corresponding $p_i$ and we can replace $q$ by $\tilde{q}:= \bigcap\limits_{j \neq i} p_i$. Then, with the help of $\tilde{q}$ we can show that the subset $Q\backslash \{p_i\}$ of size $m-1$ is also an ufg set. Note that in this case, we need not refer to $q$ to get $\tilde{q}$, we can in fact remove some arbitrary $p_i$ and can replace $q$ by $\tilde{q}:= \bigcap\limits_{j \neq i} p_i$.\\

Now we would like to do the actual proof by going through cases $i)$ and $ii)$:\\

\textcolor{red}{This is where the above error occurs:}

\fcolorbox{red}{yellow}{\parbox{\dimexpr \linewidth-2\fboxsep-2\fboxrule}{%
\textbf{Case} $\boldsymbol{i):}$ First, note that $D_{\nleq}$ is a subset of $q$. This is true, because for an element $m_i ='x_i \nleq x_j'$ we have that property (\ref{three}) is exactly the statement $q \cancel{I} m_i$  which means that $ (x_i,x_j ) \in q$, compare (\ref{mnleq}). Because $q$ is a partial order, there cannot be cycles within $D_\nleq$. (Note further that $D_\nleq$ needs not to be transitive.) In particular this implies that the items of $q$ can be totally ordered by a relation $L$ such that $L$ is a linear extension of $D_\nleq$.

Now, take some pair $(a,b) \in D_\nleq$ with the property \\

$(^*)$: that there exists no other pair $(c,d) \in D_\nleq $ with $a L c$ and $d L b$.\\

{\bfseries{Remark:}} Note that the chosen pair is not necessarily a pair in the transitive reduction of $q$, but this is actually not needed. The final goal will be to remove the pair $(a,b)$ from $q$ to obtain a candidate $\tilde{q}$. But before removing $(a,b)$ one possibly has to remove other pairs beforehand, see below. We actually chosed a pair that satisfies $(^*)$ to ensure that before removing $(a,b)$ we do not remove another distinguishing pair from $D_{\nleq}$. (This would possibly make our reduced set $Q\backslash \{p_i\}$ redundant.)

}}

 

To find a pair satisfying $(^*)$ is in fact possible because $D_\nleq$ is finite. Now we would like to remove the order $p_i$ that corresponds to the distinguishing attribute $m_i:= ' a\nleq b'$ and construct a partial order $\tilde{q}$ from the order $q$ that in contrast to $q$ does not have the pair $(a,b)$ (i.e., $(a,b) \notin \tilde{q}$, this would ensure that $Q\backslash \{p_i\} \rightarrow \{\tilde{q}\} $, at least if there are no other equivalent distinguishing attributes) and that does show with Lemma \ref{lemma_characterizing_ufg}  that $\{p_1,\ldots, p_m\}\backslash\{p_i\}$ is again an ufg set.\\

Because $\tilde{q}$ should be transitive, we cannot simply remove $(a,b)$ from $q$, because we would possibly loose transitivity. Therefore, we successively remove pairs $(c,d)$ from $q$ to get ${q}_1,\ldots , {q}_{k-1},$ and the final candidate ${q}_k=:\tilde{q}$. We proceed in the concrete way described below. By going from $q_l$ to $q_{l+1}$ we will proceed in a way that all relations ${q}_l$ are still partial orders (and in particular transitive) and in a way that it always holds that\footnote{property \ref{first_property} will be only satisfied in the case that there are no distinguishing attributes that are equivalent to $(a,b)$. In this case one has to proceed with further steps until for the first time another pair $(\tilde{a},\tilde{b})$ along with an order $p_j$ can be removed to obtain a proper subset of $Q$ with size $m-1$.  }:

\begin{align}
\{p_1,\ldots, p_m\} \backslash \{p_i\} \rightarrow \tilde{q} &\mbox{ , because in the end } (a,b) \notin \tilde{q} ) & \label{first_property}\\ 
\forall l < k : & \{p_1,\ldots, p_m\} \backslash \{p_i\}\; \cancel{\rightarrow} \; \tilde{q}_l \mbox{ only because } (a,b) \in \tilde{q}_l & \label{second_property}\\
&\mbox{ (and also because of possible equivalent distinguishing} \nonumber \\
& \mbox{ attributes) } \nonumber \\
\forall l: \forall (c,d) \in D_\nleq \backslash \{ (a,b) \}:& (c,d) \in \tilde{q}_l \mbox{ (this ensures that in the end all $p_j$ with $i\neq j $} & \label{third_property}\\
&\mbox{ are not redundant in implying $\tilde{q}_k$.)} \nonumber 
\end{align}

The concrete way to proceed is the following:

First note that $(a,b) \in q$ because $D_{\nleq}$ is a subset of $q$. Now we look at every path $(a, z_1),(z_1,z_2), \ldots ,(z_l,b)$ of neighboured elements (w.r.t. $q$) from $a$ to $b$.
Because every $p_j$ with $j\neq i$ has the property $(a,b) \notin p_j$ there exists (for some arbittrary choosen $j\neq i$) one $(z_t,z_{t+1}) $ with $(z_t,z_{t+1}) \notin p_j$. Thus, removing the pair $(z_t,z_{t+1}) $ from $\tilde{q}_l$ will not destroy property (\ref{second_property}) and in fact will lead to a transitive relation.

Additionally, property (\ref{third_property}) will also be kept because all $(z_t,z_{t+1}) $ are no distinguishing attributes from $D_\nleq$ because  $(z_t,z_{t+1}) $ is between the pair $(a,b)$ w.r.t. relation $L$ (cf., property $(^*)$.)

After having deleted one pair $(z_t,z_{t+1})$ for every path from $a$ to $b$ one can now delete also the pair $(a,b)$ without loosing transitivity and one will get the final $\tilde{q}_k$ that has also property (\ref{first_property}). Together with property (\ref{third_property}) this $\tilde{q}_k$ in fact shows that $\{p_1,\ldots,p_m\}\backslash \{p_i\}$  is an ufg set of size $m-1$.\\

\textbf{In the case} $\boldsymbol{ii)}$ one can simply remove some arbitrary $p_i$ from $\{p_1,\ldots ,p_m\}$ to get $Q:=\{p_1, \ldots,p_m\}\backslash \{p_i\}$. That every $p_i$ can be choosen follows from the fact that (within $D$) each $p_i$ with $i \in \{1, \ldots, m\}$ differs from all others only by the existence of non-pairs. Thus, with the partial order $\tilde{q}:= \bigcap\limits_{j\neq i} p_j$ we have:

\begin{align*}
Q\rightarrow \{\tilde{q}\} & \mbox{ (because $\tilde{q}$ is the intersection of the orders in } Q \mbox{ (and as such it is also a}\\
&\mbox{ subset of the union of the orders in $Q$)}\\
\forall R\subsetneq Q: R \cancel{\rightarrow } \{\tilde{q} \} & \mbox{ because every $p_j$ with $j\neq i$   has some distinguishing attribute}\\
& \mbox{ $m_i= 'a\leq b'$ such that $\bigcap\limits_{l \neq j, l\neq i} p_l \ni (a,b)$ but $(a,b) \notin \tilde{q}$ (because $(a,b)\notin p_j$)}.
\end{align*}

Case i) and ii) show that the reduced set $Q$ is in fact an ufg set of size $m-1$.\\


\end{proof}

    \bibliography{literature_notes.bib}

\begin{thebibliography}{5}
\providecommand{\natexlab}[1]{#1}
\providecommand{\url}[1]{\texttt{#1}}
\expandafter\ifx\csname urlstyle\endcsname\relax
  \providecommand{\doi}[1]{doi: #1}\else
  \providecommand{\doi}{doi: \begingroup \urlstyle{rm}\Url}\fi

\bibitem[Bastide et~al.(2000)Bastide, Pasquier, Taouil, Stumme, and
  Lakhal]{bastide00}
Yves Bastide, Nicolas Pasquier, Rafik Taouil, Gerd Stumme, and Lotfi Lakhal.
\newblock Mining minimal non-redundant association rules using frequent closed
  itemsets.
\newblock In John Lloyd, Veronica Dahl, Ulrich Furbach, Manfred Kerber,
  Kung-Kiu Lau, Catuscia Palamidessi, Lu{\'i}s~Moniz Pereira, Yehoshua Sagiv,
  and Peter Stuckey, editors, \emph{Computational Logic --- CL 2000}, pages
  972--986. Springer, 2000.

\bibitem[Blocher et~al.(2022)Blocher, Schollmeyer, and Jansen]{bsj2022}
Hannah Blocher, Georg Schollmeyer, and Christoph Jansen.
\newblock Statistical models for partial orders based on data depth
  and formal concept analysis.
\newblock In Davide Ciucci, Ines Couso, Jesus Medina, Dominik Slezak, Davide
  Petturiti, Bernadette Bouchon-Meunier, and Ronald~R. Yager, editors,
  \emph{Information Processing and Management of Uncertainty in Knowledge-Based
  Systems}, pages 17--30. Springer, 2022.

\bibitem[Blocher et~al.(2023)Blocher, Schollmeyer, Christoph, and
  Nalenz]{Blocher_isipta}
Hannah Blocher, Georg Schollmeyer, Jansen Christoph, and Malte Nalenz.
\newblock Depth functions for partial orders with a descriptive analysis of
  machine learning algorithms.
\newblock \emph{Submitted to ISIPTA 2023}, 2023.

\bibitem[Ganter and Wille(2012)]{ganter2012formal}
Bernhard Ganter and Rudolf Wille.
\newblock \emph{Formal Concept Analysis: Mathematical Foundations}.
\newblock Springer Science \& Business Media, 2012.

\bibitem[Priss(2006)]{priss2006formal}
Uta Priss.
\newblock Formal concept analysis in information science.
\newblock \emph{Annual Review Information Science Technology}, 40\penalty0
  (1):\penalty0 521--543, 2006.

\end{thebibliography}
\end{document}